\documentclass{article}
\usepackage{xcolor}
\newcommand{\tagcolor}[1]{\textcolor{blue}{#1}}
\usepackage{tcolorbox}

\PassOptionsToPackage{numbers, compress}{natbib}

\usepackage[preprint]{neurips_2024}

\usepackage[utf8]{inputenc} 
\usepackage[T1]{fontenc}    
\usepackage{hyperref}       
\usepackage{url}            
\usepackage{booktabs}       
\usepackage{amsfonts}       
\usepackage{nicefrac}       
\usepackage{microtype}      
\usepackage{xcolor}         

\usepackage{amsmath}
\usepackage{amssymb}
\usepackage{mathtools}
\usepackage{amsthm}
\usepackage{mathrsfs}
\usepackage{enumitem}
\usepackage{subfig}
\usepackage{multirow}
\usepackage{graphicx} 
\usepackage{comment}
\usepackage{listings}
\usepackage{caption}
\usepackage{makecell}

\title{Learning from Failures in Multi-Attempt Reinforcement Learning}

\author{
    Stephen Chung\thanks{Equal contribution. Correspondence to: \texttt{stephen.chung@dualityrl.com}.} \\    
    DualityRL \\
    \And
    Wenyu Du$^*$\\    
    DualityRL \\  
    \And
    Jie Fu \\
    Shanghai AI Lab \\
}

\begin{document}

\maketitle
\setcounter{footnote}{0}
\begin{abstract}
Recent advancements in reinforcement learning (RL) for large language models (LLMs), exemplified by DeepSeek R1, have shown that even a simple question-answering task can substantially improve an LLM's reasoning capabilities. In this work, we extend this approach by modifying the task into a multi-attempt setting. Instead of generating a single response per question, the model is given multiple attempts, with feedback provided after incorrect responses. The multi-attempt task encourages the model to refine its previous attempts and improve search efficiency. Experimental results show that even a small LLM trained on a multi-attempt task achieves significantly higher accuracy when evaluated with more attempts, improving from 45.6\% with 1 attempt to 52.5\% with 2 attempts on the math benchmark. In contrast, the same LLM trained on a standard single-turn task exhibits only a marginal improvement, increasing from 42.3\% to 43.2\% when given more attempts during evaluation. The results indicate that, compared to the standard single-turn task, an LLM trained on a multi-attempt task achieves slightly better performance on math benchmarks while also learning to refine its responses more effectively based on user feedback.
\end{abstract}

\section{Introduction}
Recent advancements in large-scale post-training of reinforcement learning (RL) for large language models (LLMs) have demonstrated a promising approach to enhancing their reasoning capabilities. These improvements have led to emergent abilities such as self-correction and self-refinement~\citep{o1,guo2025deepseek}. Most existing methods rely on single-turn tasks, where the model receives a reward based on the correctness of its single response to a question. However, single-turn tasks may be inefficient due to sparse rewards, and they do not require the LLM to learn how to respond to user feedback. In this work, we propose a simple yet effective multi-turn task that enables LLMs to learn reasoning through RL.\footnote{Full code is available at \url{https://github.com/DualityRL/multi-attempt}}.

\begin{figure}[h]
    \centering
    \includegraphics[width=0.95\textwidth]{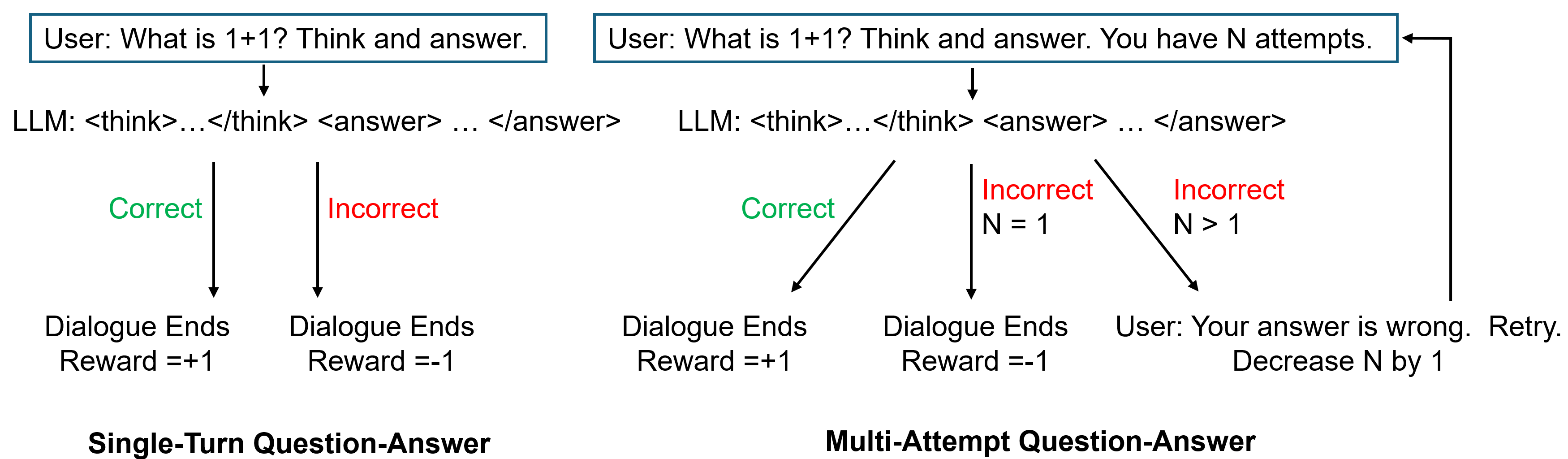}
    \caption{Illustration of the multi-attempt question-answer task. We extend the single-turn question-answer task from DeepSeek R1 to a multi-attempt setting, enabling iterative refinement.}
    \label{fig:illust}
\end{figure}

Instead of requiring the LLM to provide a single response to a given question, we propose a \emph{multi-attempt} task that allows the LLM to generate multiple responses based on feedback. Specifically, we first randomly sample $N$ as the number of remaining attempts for each question. The model initially generates a response to a given question as usual. If the response is correct or there are no remaining attempts (i.e., $N\leq1$), the dialogue ends. However, if the response is incorrect and there are remaining attempts (i.e., $N>1$), we provide feedback indicating that the answer is incorrect and prompt the LLM to try again, while decrementing the remaining attempts $N$ by 1. An illustration is shown in Figure~\ref{fig:illust}, and an example dialogue from a trained LLM is shown in Figure~\ref{fig:example_dialogue}.

\begin{figure}[h]
    \centering
    \includegraphics[width=0.65\textwidth]{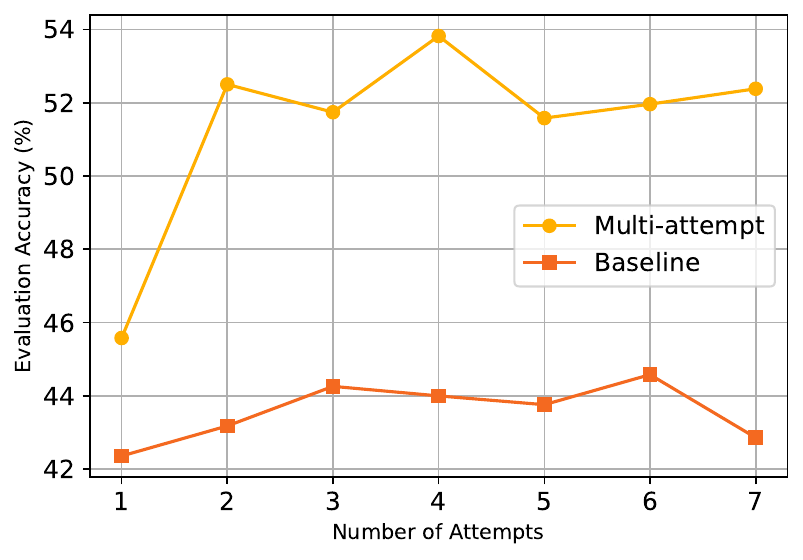}
    \caption{Evaluation accuracy as a function of the number of allowed attempts during evaluation, averaged across five benchmarks: AIME 2024, MATH 500, AMC 2023, Minerva Math, and OlympiadBench. Both LLMs are based on Qwen 2.5 Math 1.5B and fine-tuned via RL on a small math dataset in either multi-attempt tasks or single-turn tasks (baseline).}
    \label{fig:attempt_comparison}
\end{figure}

Our training pipeline is simple and applies standard RL to the multi-attempt task on a math problem dataset, largely similar to how DeepSeek R1 Zero~\citep{guo2025deepseek} is trained. In the multi-attempt task, a reward of +1 is given if the answer is correct in any attempt, -0.5 if the answer is incorrect but in the correct format, and -1 otherwise. We use standard Proximal Policy Optimization (PPO)~\citep{schulman2017proximal} as the training algorithm.

In addition, the multi-attempt task enables the LLM to learn how to respond to user feedback through RL, rather than merely answering the initial question. In most current LLMs, the ability to respond to user feedback is generally trained via RLHF~\citep{ouyang2022training} or supervised fine-tuning (SFT). Compared to RLHF or SFT, learning to respond to user feedback through pure RL may lead to more interesting emergent capabilities, such as significantly improved solution refinement, as demonstrated by our experimental results.

An interesting phenomenon observed in DeepSeek R1 Zero is the emergence of the \emph{Aha Moment}, which refers to the model's ability to recognize its mistakes (self-verification) and correct its answers (self-refinement). In DeepSeek R1 Zero, both self-verification and self-refinement capabilities emerge spontaneously within the model. Our multi-attempt approach facilitates the emergence of self-refinement, as the model has a strong incentive to refine its previous failed attempts (e.g., it does not receive any reward for repeating the same failed attempt). As such, multi-attempt tasks may enable the Aha Moment to emerge more easily, as the model is explicitly trained for self-refinement.

Our experiments show that even a small LLM, such as a 1.5B model, can effectively learn self-refinement capabilities. As illustrated in Figure \ref{fig:attempt_comparison}, the evaluation accuracy of an LLM trained on a multi-attempt task improved from 45.6\% to 52.5\% on math benchmark when increasing the number of attempts from 1 to 2. In contrast, the same model trained on a single-turn task showed only a marginal gain, from 42.3\% to 43.2\%. We also observe that even under the standard 1-attempt evaluation, the multi-attempt LLM slightly outperforms its single-turn counterpart, highlighting the benefits of multi-attempt training. 

As RL gains popularity in LLMs for enhancing reasoning capabilities, we believe that a multi-turn setting, in general, provides a richer learning signal and better enhances capabilities compared to a single-turn setting. We foresee that more sophisticated capabilities could emerge in LLMs trained in a multi-turn setting via RL.

\begin{figure}[h]
\centering
\fbox{
\begin{minipage}{0.95\linewidth}
\textbf{Example Dialogue:}

\texttt{System:} A conversation between User and Assistant. The User asks a question, and the Assistant solves it. The Assistant first thinks about the reasoning process in the mind and then provides the User with the answer. The reasoning process is enclosed within \tagcolor{<think>} \tagcolor{</think>} and the answer is enclosed within \tagcolor{<answer>} \tagcolor{</answer>} tags, respectively, i.e., \tagcolor{<think>} reasoning process here \tagcolor{</think>} \tagcolor{<answer>} answer here \tagcolor{</answer>}. 

\texttt{User:} \textbf{You have 2 attempts to answer the question.} Your final answer will be extracted automatically by the boxed tag. This is the problem:

\begin{tcolorbox}
What is the value of \( c \) if \( x\cdot(3x+1)<c \) if and only if \( x\in \left(-\frac{7}{3},2\right) \)?
\end{tcolorbox}

\texttt{Assistant:} \tagcolor{<think>}

To solve this problem, we need to determine the value of \( c \) such that the inequality \( x \cdot (3x + 1) < c \) holds true if and only if \( x \in \left( -\frac{7}{3}, 2 \right) \). Let's break it down step-by-step:

1. \textit{Understanding the Given Condition:}
   - We need to find \( c \) such that \( x \cdot (3x + 1) < c \) within the given interval.

2-4. (Omitted for brevity)

5. \textit{Conclusion:}
   - Both \(\frac{140}{3}\) and 14 are needed to be evaluated accurately. Checking both values:
     - \(\frac{140}{3} \approx 46.67\)
     - \(14\) is less than \(\frac{140}{3}\).

Thus, the highest value that fits our condition is \( c = \frac{140}{3} \).

\tagcolor{</think>}

\tagcolor{<answer>} \(\boxed{\frac{140}{3}}\) \tagcolor{</answer>}

\texttt{User:} \textbf{Your previous answer is wrong. Try alternatives and refine your answer. You have 1 attempt left.}

\texttt{Assistant:} \tagcolor{<think>}

I apologize for the mistake. Let's re-evaluate the problem step-by-step after rechecking our previous work.

1. \textit{Understanding the Given Condition:}
   - We need to find \( c \) such that \( x \cdot (3x + 1) < c \) within the given interval.

2-4. (Omitted for brevity)

5. \textit{Conclusion:}
   - The correct maximum value is actually 14 from \( x = 2 \).

Thus, the highest value that fits our condition is \( c = 14 \).

\tagcolor{</think>}

\tagcolor{<answer>} \(\boxed{14}\) \tagcolor{</answer>}
\end{minipage}
}
\caption{An example of a multi-attempt dialogue ($N=2$) from a fine-tuned LLM, where the LLM makes a mistake on the first attempt but learns to correct it in the second attempt.}
\label{fig:example_dialogue}
\end{figure}
\clearpage

\section{Approach}

In a \emph{single-turn} task, a question is sampled from a dataset, and the LLM generates a response to the question. Concretely, let the dataset be denoted as $
\mathcal{D} = \{(x_i, y^*_i)\}_{i=1}^{N}$ where \( x_i \) is question \( i \), \( y^*_i \) is the ground-truth answer for question \( x_i \), and \( N \) is the size of the dataset. Denoting the policy of the model as \( \pi_\theta(\cdot|x_i) \), parameterized by \( \theta \), we sample a response \( a_i \) to question \( x_i \). The objective is to maximize:
\begin{equation}
    J(\theta) = \mathbb{E}_{x, y^* \sim \mathcal{D}}[R(a, y^*)], 
\end{equation}
where $a \sim \pi_\theta(\cdot|x)$, and \( R \) is the reward function, such as a binary function that gives \( 1 \) if the extracted answer from \( a \) is the same as the ground-truth answer \( y^* \) and \( -1 \) otherwise.

In a more general \emph{multi-turn} task, we allow the dialogue to continue after the first response. Concretely, we denote \( x_t \) and \( a_t \) as the prompt and model response at turn \( t \). The initial prompt \( x_0 \) is randomly sampled from the questions in the dataset \( \mathcal{D} \). To generate the subsequent prompt \( x_t \), we define the transition function $x_t = g(x_0, a_0, x_1, a_1, \dots, x_{t-1}, a_{t-1})$ which determines the next prompt based on previous prompts and responses or whether to terminate the episode. Thus, the objective in multi-turn task becomes:
\begin{equation}
    J(\theta) = \mathbb{E}_{x_0, y^* \sim \mathcal{D}} \left[ \sum_{t=0}^\infty R(a_t, y^*) \right],  \label{eq:mul}
\end{equation}
where $ a_t \sim \pi_\theta(\cdot | x_0, a_0, x_1, a_1, \dots, x_{t-1})$, that is, the response conditioned on all previous prompts and responses.

In the proposed \emph{multi-attempt} task, we first determine the number of attempts $N=1, 2, ...$ and adopt a specific transition function \( g \), given by:

\begin{equation}
    \small
    g(x_0, a_0, \dots, x_{t-1}, a_{t-1}) =
    \begin{cases} 
        \text{Dialogue termination,} & \text{if } t > N \text{ or } \text{Extract}(a_{t-1}) = y^*, \\[4pt]
        \text{Prompt indicating the remaining attempts} \\ 
        \quad \text{and requesting another attempt,} & \text{otherwise}.
    \end{cases}
\end{equation}

In other words, if the model answers correctly or the attempts are exhausted, the dialogue ends; otherwise, the model is prompted for another attempt. An illustration is shown in Figure~\ref{fig:illust}, and a dialogue example from a trained model is shown in Figure~\ref{fig:example_dialogue}. 

In our experiment, we uniformly sample \( N \) from \( \{1, \dots, M\} \) for each question, where \( M \) is the maximum number of allowed attempts (e.g., \( M = 5 \)). This random sampling encourages the model to learn more diverse behaviors based on the number of available attempts. For example, if a question is sampled with \( N = 1 \) (as in the evaluation phase, where we assume the ground-truth answer is unknown), the model should learn to respond carefully and greedily, as it has only one chance. However, when \( N \) is large, the model can explore more freely in early attempts, knowing that additional chances remain.

Our reward function $R$ is defined as follows:
\begin{itemize}
    \item \( +1 \) if the model answers correctly in any attempt,
    \item \( -0.5 \) if the model provides a response in the correct format but with an incorrect answer,
    \item \( -1 \) otherwise.
\end{itemize}

We do not discount rewards based on the number of attempts used, ensuring that the model is encouraged to explore freely in early attempts without incurring a penalty.

Note that the objective in the multi-turn task (\ref{eq:mul}) can be directly optimized via standard RL algorithms such as PPO. As the focus of this work is on the task used to fine-tune LLMs, we refer readers to previous works on finetuning LLM with RL \citep{guo2025deepseek} for training details.

\section{Related Work}

\textbf{Learning to Plan in RL.} Our approach is inspired by the Thinker\citep{chung2023thinker} algorithm, a model-based RL method that enables an agent to explore different alternatives within a learned world model before taking a real action. However, instead of learning a world model, we directly augment the environment to allow multiple attempts.

\textbf{Multi-turn Tasks for LLMs.} A similar approach is SCoRe\citep{kumar2024training}, which also trains an LLM on multi-attempt math tasks via RL. However, their method does not use the ground-truth reward to verify whether the previous attempt was successful. As a result, it requires careful calibration and a two-stage training process to prevent policy collapse, since the LLM has a strong incentive to replicate its first attempt. 
Other multi-turn studies, such as \citep{zhou2024archer, shani2024multi, shi2024direct,jain2025multiturncodegenerationsinglestep}, primarily focus on designing the RL algorithms for optimizing multi-turn tasks rather than designing a general RL environment to enhance the reasoning capabilities of LLMs. While these algorithms could be applied in our proposed multi-attempt task, we adopt standard PPO in this work for simplicity.

\textbf{Self-Correction with External Tools for LLMs.} Numerous prior studies have explored self-correction mechanisms in LLMs \citep{kamoi2024can}. Reflexion \citep{shinn2023reflexion} enables LLMs to engage in self-reflection based on external or internal feedback, guiding future trials. CRITIC \citep{gou2023critic} facilitates interactions with external tools to provide immediate feedback. \citep{chenteaching} allows LLMs to execute code for debugging, while FLARE \citep{jiang2023active} leverages a retriever to estimate the probability of output sentences, enabling fact-checking. In contrast to these works, our work extends the single-turn question-answering task used in DeepSeek R1 \citep{guo2025deepseek} into a multi-attempt setting based on a similar RL pipeline to enhance general reasoning capabilities of LLM.

\textbf{Learning from Historical Logs.} Many studies\citep{dusek-jurcicek-2016-context,snell2022contextawarelanguagemodelinggoaloriented,zheng2022hiesqlhistoryinformationenhanced} explore leveraging past relevant logs to enhance current predictions. For instance, \cite{zheng2022hiesqlhistoryinformationenhanced} retrieves the past user queries and their predicted SQL from historical logs, providing additional context for generating current SQL. In contrast, our approach utilizes self-generated past attempts as failure examples, encouraging the model to learn from previous mistakes to improve subsequent attempts.

\section{Experiments}
We fine-tune a small pretrained model, namely Qwen 2.5 Math 1.5B, on 8K math questions provided in \citep{zeng2025simplerl}. We use PPO with a discount rate of \( \gamma = 1 \), lambda \( \lambda = 0.99 \), and a small KL divergence coefficient of \( 0.01 \). The LLM is trained for 160 episodes, generating a single sample per question in each episode, totaling \( 160 \times 8\text{K} = 1.28\text{M} \) training samples. Our code, modified from \citep{zeng2025simplerl}, is publicly available.

In the multi-attempt experiment, we set \( M = 5 \), meaning the number of attempts is uniformly sampled from \(\{1, 2, ..., 5\}\) for each question. In the baseline, we use the standard single-turn setting, where the agent receives a reward of \( +1 \) for a correct answer, \( -0.5 \) for an incorrect answer in the correct format, and \( -1 \) otherwise.

\begin{figure}[h]
    \centering
    \subfloat[Training Reward]{%
        \includegraphics[width=0.45\textwidth]{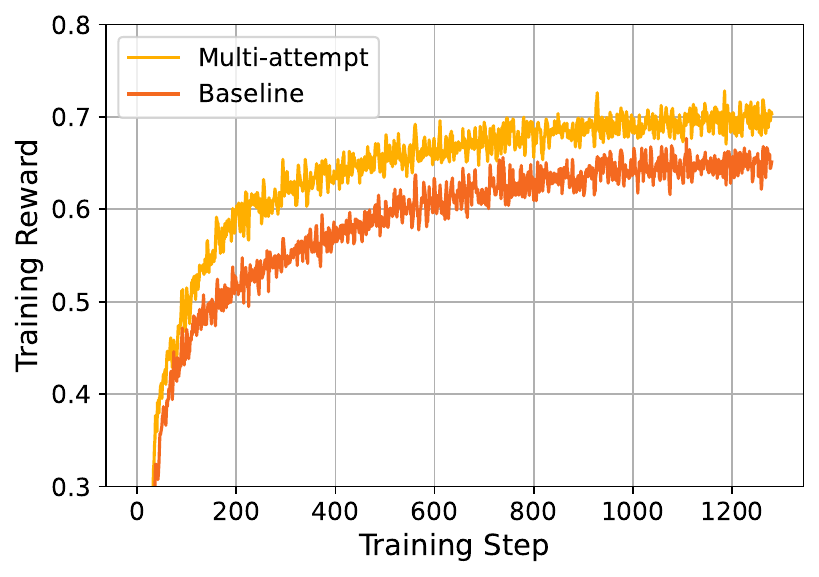}
        \label{fig:tr}
    }
    \hfill
    \subfloat[Average Evaluation Accuracy]{%
        \includegraphics[width=0.45\textwidth]{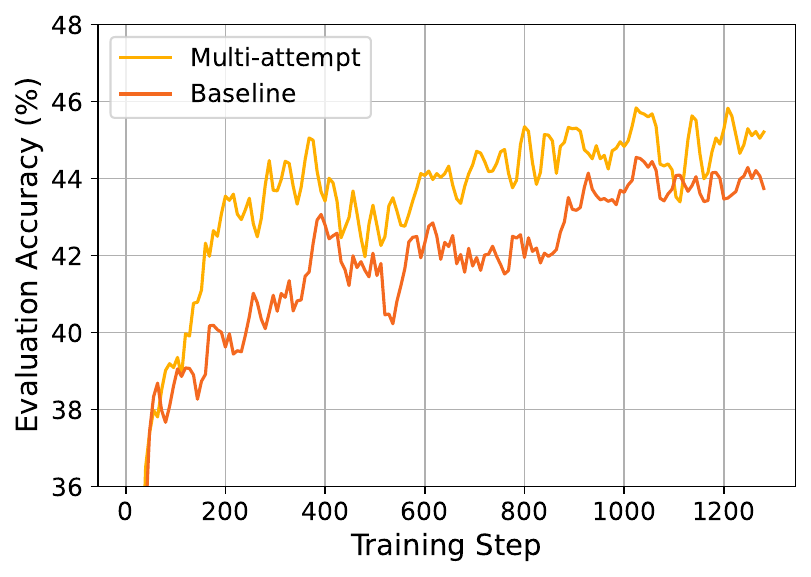}
        \label{fig:lc}
    }

    \caption{Training and evaluation performance of the LLMs. (a) Training reward as a function of training steps. (b) Average evaluation accuracy across five benchmarks as a function of training steps, evaluated under the standard single-attempt setting.}
    \label{fig:training}
\end{figure}

The reward during training is shown in Figure \ref{fig:tr}. As expected, the multi-attempt LLM collects more rewards than the baseline LLM, as the multi-attempt setting is inherently easier to solve. The learning curve, which represents the average evaluation performance across five benchmarks (AIME 2024, MATH 500, AMC 2023, Minerva Math, and OlympiadBench), is shown in Figure \ref{fig:lc}. For a fair comparison, we evaluate both LLMs under the standard single-attempt setting. The results indicate that the multi-attempt LLM slightly outperforms the baseline, suggesting that multi-attempt training may even benefit standard single-attempt evaluations. We hypothesize that this improvement stems from the broader and more efficient exploration facilitated by multi-attempt tasks, leading to more effective reinforcement learning.

\begin{table}[h]
    \centering
    \renewcommand{\arraystretch}{1.2} 
    \setlength{\tabcolsep}{6pt} 
    \begin{tabular}{lcccccc}
        \hline
         & \textbf{Avg.} &  & \textbf{AIME} & \textbf{AMC} & \textbf{Minerva} & \textbf{Olympiad} \\
         \textbf{Method} & \textbf{Accuracy} & \textbf{MATH500} & \textbf{2024} & \textbf{2023} & \textbf{Math} & \textbf{Bench} \\
        \hline
        Multi-Attempt  & 45.4  & 73.4  & 20.0  & 65.0  & 35.3  & 33.9 \\
        Baseline       & 43.5  & 75.4  & 13.3  & 55.0  & 35.3  & 37.5 \\
        \hline
    \end{tabular}
    \vspace{5pt}
    \caption{Comparison of evaluation accuracy across multiple benchmarks. }
    \label{tab:benchmark_comparison}
\end{table}

\begin{figure}[h]
    \centering
    \subfloat[AIME 2024]{%
        \includegraphics[width=0.47\textwidth]{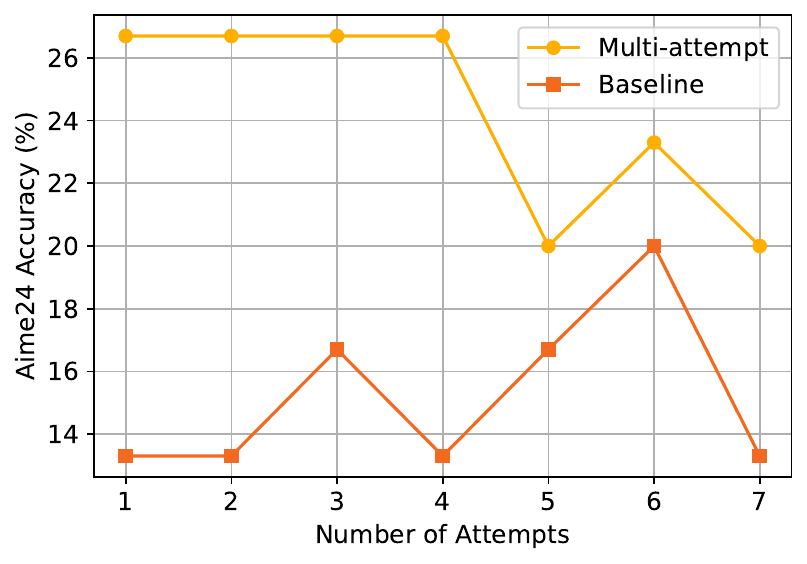}
        \label{fig:aime24}
    }
    \hfill
    \subfloat[MATH500]{%
        \includegraphics[width=0.47\textwidth]{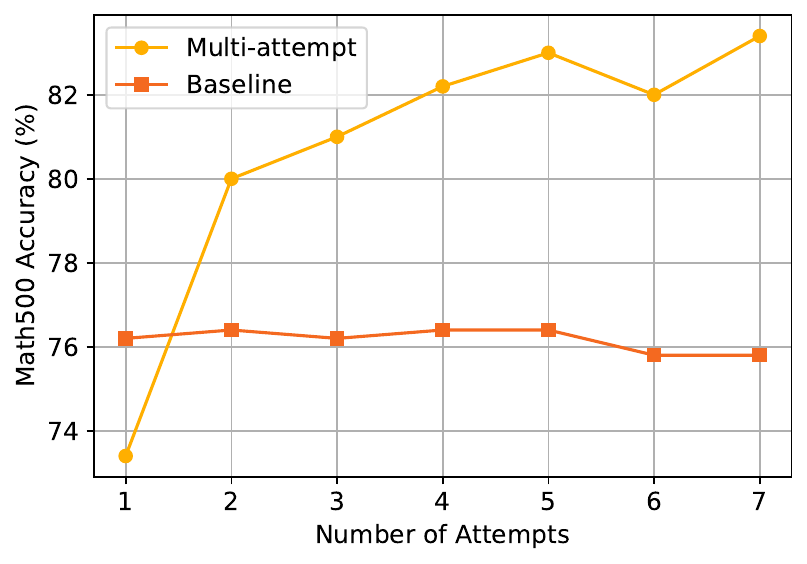}
        \label{fig:math500}
    }
    
    \vspace{10pt} 

    \subfloat[AMC 2023]{%
        \includegraphics[width=0.47\textwidth]{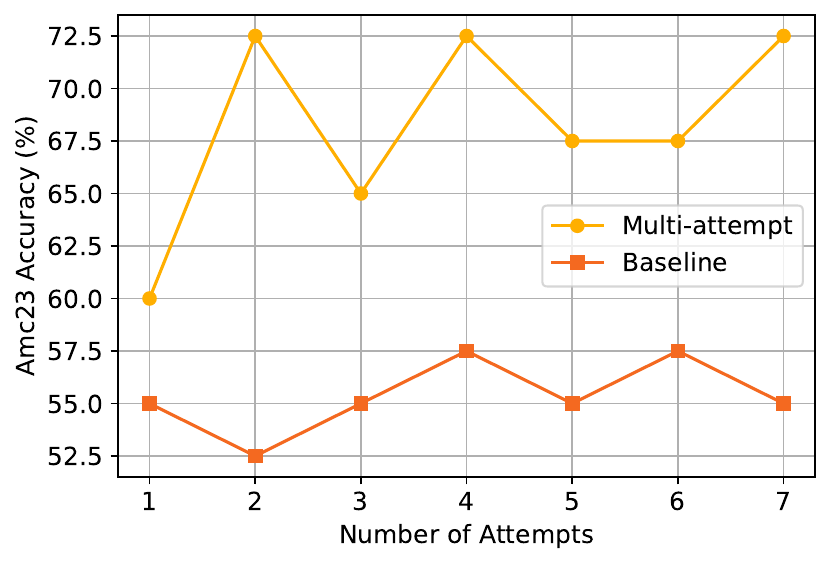}
        \label{fig:amc23}
    }
    \hfill
    \subfloat[Minerva Math]{%
        \includegraphics[width=0.47\textwidth]{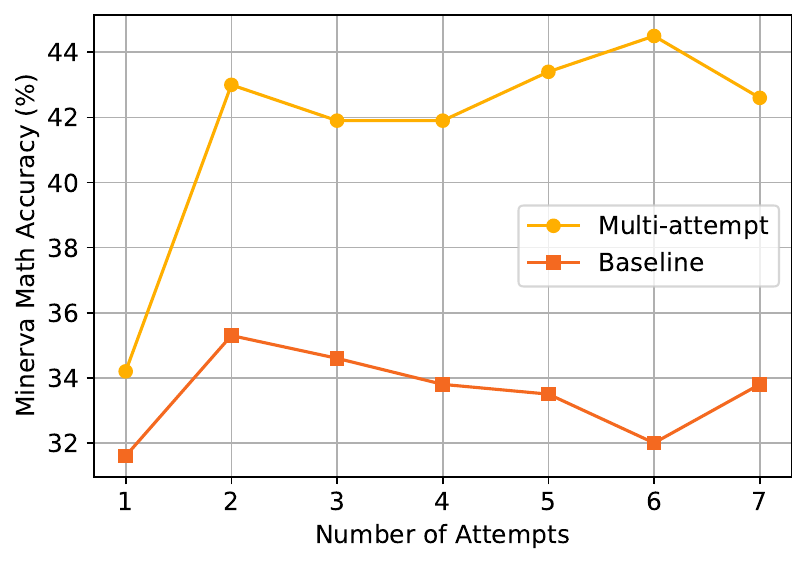}
        \label{fig:minerva_math}
    }

    \vspace{10pt} 

    \subfloat[OlympiadBench]{%
        \includegraphics[width=0.47\textwidth]{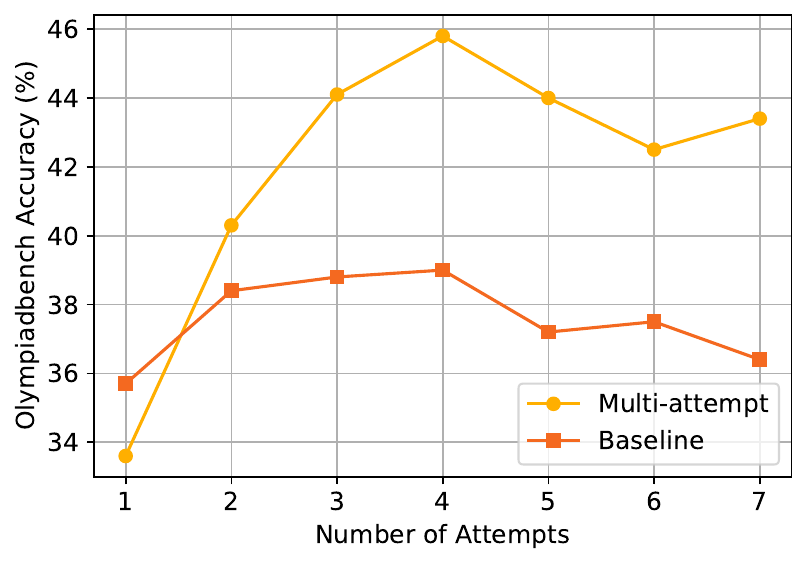}
        \label{fig:olympiadbench}
    }

    \caption{Evaluation accuracy as a function of the number of allowed attempts during evaluation on individual benchmarks.}
    \label{fig:attempt_sub}
\end{figure}

Another important metric to evaluate is how effectively the LLM refines its search based on previous failed attempts. We assess this by allowing multiple attempts in the math benchmark, with the average results shown in Figure \ref{fig:attempt_comparison} and individual results presented in Figure \ref{fig:attempt_sub}. The results indicate that when trained in a multi-attempt setting, the LLM effectively leverages prior failed attempts, significantly improving performance from 45.58\% to 53.82\% as the number of attempts increases from 1 to 4. In contrast, the baseline model sees only marginal benefits from multiple attempts, suggesting that its refinement capability is weaker compared to an LLM explicitly trained in a multi-attempt setting. This ability to iteratively refine responses based on user feedback could be particularly valuable in domains requiring adaptive reasoning, such as assisting users in complex code generation.

\section{Conclusion}
In this work, we extend the question-answering task used in DeepSeek R1 by introducing a multi-attempt mechanism. Our experiments demonstrate that while the multi-attempt task provides a modest improvement in base performance on math evaluation benchmarks, it significantly enhances the model’s ability to correct mistakes based on user feedback. We envision that further improvement to the task environment—such as incorporating more nuanced and detailed feedback or introducing auxiliary tasks—could foster different capabilities in LLMs and present valuable directions for future exploration.

\section*{Acknowledgment}
Stephen Chung completed this work at the University of Cambridge, while Wenyu Du completed this work at the University of Hong Kong.

\clearpage

\bibliographystyle{unsrtnat}
\bibliography{citation}
\clearpage

\end{document}